\documentclass{article}

     \PassOptionsToPackage{numbers, compress}{natbib}

\usepackage[final]{neurips_2021}

\usepackage[utf8]{inputenc} 
\usepackage[T1]{fontenc}    
\usepackage{hyperref}       
\usepackage{url}            
\usepackage{booktabs}       
\usepackage{amsfonts}       
\usepackage{nicefrac}       
\usepackage{microtype}      
\usepackage[ruled,vlined]{algorithm2e}

\usepackage{amsmath,amsfonts,bm}

\def\eqref#1{equation~\ref{#1}}

\def\1{\bm{1}}

\DeclareMathAlphabet{\mathsfit}{\encodingdefault}{\sfdefault}{m}{sl}
\SetMathAlphabet{\mathsfit}{bold}{\encodingdefault}{\sfdefault}{bx}{n}

\def\gA{{\mathcal{A}}}

\def\gD{{\mathcal{D}}}

\def\gF{{\mathcal{F}}}
\def\gG{{\mathcal{G}}}

\def\gN{{\mathcal{N}}}
\def\gO{{\mathcal{O}}}

\def\gR{{\mathcal{R}}}
\def\gS{{\mathcal{S}}}
\def\gT{{\mathcal{T}}}
\def\gU{{\mathcal{U}}}

\def\gY{{\mathcal{Y}}}
\def\gZ{{\mathcal{Z}}}

\def\sP{{\mathbb{P}}}

\newcommand{\E}{\mathbb{E}}

\DeclareMathOperator*{\argmin}{arg\,min}

\usepackage{amsmath,amsfonts,bm}

\def\EX{{\mathbb{E}}}

\makeatletter
\newcommand{\subalign}[1]{%
  \vcenter{%
    \Let@ \restore@math@cr \default@tag
    \baselineskip\fontdimen10 \scriptfont\tw@
    \advance\baselineskip\fontdimen12 \scriptfont\tw@
    \lineskip\thr@@\fontdimen8 \scriptfont\thr@@
    \lineskiplimit\lineskip
    \ialign{\hfil$\m@th\scriptstyle##$&$\m@th\scriptstyle{}##$\hfil\crcr
      #1\crcr
    }%
  }%
}
\makeatother

\usepackage{appendix}      
\usepackage{graphicx} 
\usepackage{subfigure} 
\usepackage{wrapfig}
\usepackage{amsmath}
\usepackage{tabularx}
\usepackage{multirow}

\usepackage{lipsum}

\title{Follow the Object: Curriculum Learning for Manipulation Tasks with Imagined Goals}

\author{%
  Ozsel Kilinc \\
  WMG\\
  University of Warwick\\
  Coventry, UK CV4 7AL \\
  \texttt{ozsel.kilinc@warwick.ac.uk} \\
   \And
  Giovanni Montana \\
  WMG\\
  University of Warwick\\
  Coventry, UK CV4 7AL \\
  \texttt{g.montana@warwick.ac.uk} \\
}

\begin{document}
\maketitle

\begin{abstract}

Learning robot manipulation through deep reinforcement learning in environments with sparse rewards is a challenging task. In this paper we address this problem by introducing a notion of imaginary object goals. For a given manipulation task, the object of interest is first trained to reach a desired target position on its own, without being manipulated, through physically realistic simulations. The object policy is then leveraged to build a predictive model of plausible object trajectories providing the robot with a curriculum of incrementally more difficult object goals to reach during training. The proposed algorithm, Follow the Object (FO), has been evaluated on 7 MuJoCo environments requiring increasing degree of exploration, and has achieved higher success rates compared to alternative algorithms. In particularly challenging learning scenarios, e.g. where the object's initial and target positions are far apart, our approach can still learn a policy whereas competing methods currently fail.

\end{abstract}


\section{Introduction}

Reinforcement Learning (RL) aims to solve sequential decision making problems where the decision maker seeks to find the optimal actions that maximise a feedback signal \citep{sutton1998reinforcement}. Robotic problems are characterised by the need to make sequential decisions, and recent advances in this domain have been made through deep RL. Typical applications involve locomotion \citep{LillicrapHPHETS15, schulman2015trust, mnih2016asynchronous} and manipulation tasks \citep{fu2016one, GuHLL17},  such as grasping \citep{levine2016end, PopovHLHBVLTER17}, stacking \citep{NairMAZA18} and dexterous hand manipulation \citep{zhu2018dexterous, andrychowicz2018learning}. Since the large majority of today's RL algorithms are still sample inefficient, learning from trial and error in the real world is often unfeasible, and it is commonplace to train RL policies for continuous control using a simulated environment, e.g. through a physics engine such as MuJoCo \citep{todorov2012mujoco}. 

Traditional RL approaches for robotics have required manually designed and problem-specific reward functions providing a smoothly varying feedback signal for every visited state \citep{NgHR99, RandlovA98, PopovHLHBVLTER17, GuHLL17}.  Recently, there has been increasing evidence that similar or even superior policies for robotics tasks can be obtained  using sparse rewards, i.e. when non-zero rewards are given to the agent only when the task is successfully completed \citep{VecerikHSWPPHRL17}. Although exploration becomes much harder in this setting, one particular approach - Hindsight Experience Replay (HER) \citep{AndrychowiczCRS17} - has been proven to enable learning. 

Despite such progress, contradicting claims have been reported in the literature about the empirical performance of HER even for simple robotics tasks. For instance, in \textit{PickAndPlace}, HER has been found to fail in some cases \cite{NairMAZA18} or to succeed in other cases, provided that some of the training episodes are started from `easy' states, e.g. when the object has already been grasped by the robot \cite{AndrychowiczCRS17}. Other studies have indicated that initial grasping is not a prerequisite to success as long as the object's target position is `easy', e.g. the target position is on the table in at least some training episodes \cite{PlappertHER2}.  Overall, the currently available experimental evidence confirms what intuition alone would suggest: starting from 
`easy' configurations generally leads to successful roll-outs whereas more `difficult' initial configuration may significantly hinder the learning ability \citep{NairMAZA18}.

Inspired by these findings, we set out to investigate mechanisms to further improve upon the performance of HER on manipulation tasks that are currently difficult to learn, e.g. when the object's target position is far apart from its starting position. These situations require a significantly higher degree of exploration which makes learning with only sparse rewards  particularly challenging. The approach we propose stems from recognising the importance of learning over a curriculum of progressively more difficult goals. Curriculum learning is based upon the  principle that mastering simple skills first can help master harder ones later \citep{BengioLCW09}. However, how to implement it on manipulation tasks with only sparse reward signals is not obvious. 

The approach proposed here relies on a mechanism to imagine object positions that are progressively more difficult for the object to reach during training. These intermediary goals guide the progression of mastering simple skills first (e.g. pushing an object to a nearby target) followed by more difficult ones. Our developments in this direction are inspired by the notion of {\it object locomotion policies} \citep{KilincSimulated2019}, where the objects involved in a manipulation task are modelled as independent agents that must learn a locomotion policy (i.e. how to move from any initial position to a final one). The authors  used the object policies to define auxiliary rewards for the main robot manipulation task at hand. 

\begin{figure}[t!]
	\begin{center}
		\centerline{\includegraphics[width=\columnwidth,trim={0cm 10.4cm 7cm 2.35cm},clip]{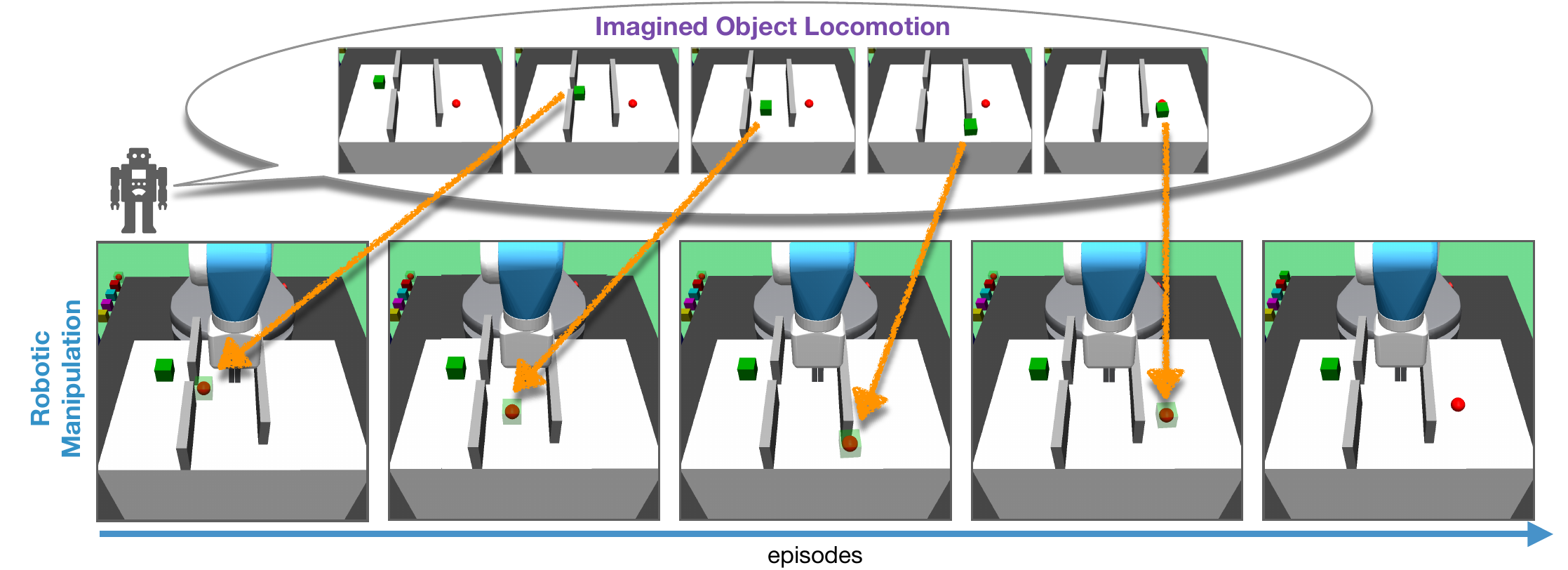}}
		\caption{An {\it object locomotion policy} allows an object (green block) to reach its final target position (red dot). This policy is used to generate trajectories of intermediary goals: imagined object positions closer to the object's initial position are easier to achieve and those closer to the original target are more difficult. The robot's learning procedure follows a curriculum: it starts with the simplest imagined goal, and progressively proceeds to a more difficult one once the robot has mastered the current difficulty level. Ultimately, the robot is able to solve the original task with only sparse rewards.}
		\label{fig:idea_vis}
	\end{center}
	\vskip -0.3 in
\end{figure}

We initially follow a similar approach as in \citep{KilincSimulated2019}, and obtain an object locomotion policy for the given task. Learning such policies successfully with only sparse rewards can be accomplished using a physics engine such as MuJoCo where objects are allowed to move freely in space, on their own, without being manipulated. Our key contribution here is to leverage the object policy to implement a curriculum learning strategy for the manipulation task. As the robot learns the task, imagined object positions are produced, each one conditional on the object's initial and target position. These imagined positions act as intermediary goals of increasing complexity. Since this complexity can be controlled precisely, the resulting learning strategy facilitates mastering simple tasks first followed by more difficult ones, and results in higher success rate overall.  The proposed approach, Follow the Object (FO), is illustrated in Figure \ref{fig:idea_vis} and described in Section \ref{sec:methodology}. Its performance  has been tested on 7 MuJoCo environments using a Fetch robotic arm with different degrees of task complexity and different starting configurations. Our empirical results and comparisons to related algorithms are presented in Section \ref{sec:experiments}, and indicate that FO is able to learn in particularly challenging scenarios whereas competing methods currently fail.

\section{Related Work} \label{relatedwork}

{\bf Imitation learning.} Imitation Learning (IL) approaches have been used to facilitate reinforcement learning of robotic manipulation tasks with sparse rewards \citep{VecerikHSWPPHRL17, HesterVPLSPHQSO18, NairMAZA18}. IL casts the sequential decision making problem as a supervised learning problem requiring either pre-collected observational data  \citep{Pomerleau88, DuanASHSSAZ17} or an interactive demonstrator \citep{RossGB11, RatliffBS07}. Demonstrations can also be used to learn a reward function \citep{NgR00, FinnLA16, HoE16}. In general, collecting demonstrations for robotics tasks can be time consuming and requires a dedicated data collection setup, e.g. virtual reality or tele-operation facility. 

{\bf Goal conditional policies.} Recently, the advantages of goal-conditioned policies for robotics problems have been discussed \citep{SchaulHGS15}. Different approaches have investigated how to automatically generate such goals. For instance, \citep{SukhbaatarLKSSF18} follow a self-play approach on reversible or resettable environments, \citep{FlorensaHGA18} employ adversarial training for robotic locomotion tasks, \citep{NairPDBLL18} use variational autoencoders for visual robotics tasks and \citep{AndrychowiczCRS17} propose an experience sampling approach, Hindsight Experience Replay (HER), to randomly draw synthetic goals from previously experienced trajectories. HER in itself may be seen as a form of implicit curriculum learning. The effectiveness of this approach has been demonstrated on a suite of challenging continuous control environments \citep{PlappertHER2}. 

{\bf Curiosity.} In these approaches, the notion of \textit{novelty} of a state provides the RL agent with an additional reward and motivation for exploring less novel states \citep{schmidhuber1991curious, burda2018large}. Curiosity has been defined in many ways, e.g. as the error in predicting the RL agent's actions by an inverse dynamics model \citep{pathak2017curiosity}, as the error in predicting the output of a fixed randomly initialised neural network \citep{burda2018exploration}, or by evaluating the novelty of a recently visited state according to its reachability from those stored in a memory \citep{savinov2018episodic}. The large majority of these works describe gaming or maze environments with discrete action spaces, but a locomotion task with continuous actions has also been considered \citep{savinov2018episodic}.

{\bf Curriculum learning.} Curriculum learning has been widely adopted into the RL framework and many researchers have sought to find a solution to the problem of how to automatically generate a way to order the goals according to difficulty. For example, \citep{tian2017elf} and \citep{wu2016training} propose to automatically adjust the difficulty of the gaming environment, \citep{czarnecki2018mix} use the policies of the intermediate agents obtained during the training, \citep{justesen2018automated} evaluate the occurrence frequencies of pre-defined events. In robotics, \citep{florensa2017reverse} propose to schedule the initial positions, \citep{eppe2019curriculum} create easier sub-goals spanning over one dimension of the target position, and \citep{fang2019curriculum} select experiences adaptively according to diversity-based curiosity.

{\bf Object motion prediction.} A body of works exists on predicting how objects behave under manipulative actions. For example, \citep{kopicki2011learning} learn a probabilistic model using a real system where a robotic arm applies random pushes to various objects, and \citep{kunze2011simulation} propose a simulation-based framework. A combination of these methods has also been proposed \citep{belter2014kinematically}. Some efforts have also been made to model the dynamics of complex scenes involving non-rigid bodies, e.g. using a particle-based simulator \citep{li2018learning}. Similarly, the object locomotion policies we use here can predict the most likely trajectory that an object has to travel to reach a given target. Unlike these works, our policies are learned irrespective of the manipulative actions exercised by the robot. Advantages and disadvantages of this approach are discussed in Section \ref{sec:evaluation}.

\section{Problem Definition and Background}

\subsection{Multi-goal RL for Robotic Manipulation}

We are concerned with solving a manipulation task: an object is presented to the robot, and has to be manipulated so as to reach a target position. In the tasks we consider, the target goal is specified by the object location and orientation, and the robot is rewarded only when it reaches its goal. We model the robot's sequential decision process as a Markov Decision Process (MDP) defined by a tuple, $M=\langle \gS, \gG, \gA, \gT, \gR, \gamma \rangle$, where $\gS$ is the set of states, $\gG$ is the set of goals, $\gA$ is the set of actions, $\gT$ is the state transition function, $\gR$ is the reward function and $\gamma \in \left[0, 1\right)$ is the discounting factor. At the beginning of an episode, the environment samples a goal $g \in \gG$.  The position of the object at time $t$ is denoted by $o_t$ and the state of the environment is $s_t \in \gS$, which includes $o_t$. We assume that, given $s_t$, we can recover $o_t$ through a known mapping, i.e. $o_t=m(s_t)$. A robot's action is controlled by a deterministic policy, i.e. $a_t=\mu_\theta(s_t,g): \gS \times \gG \rightarrow \gA$, parameterised by $\theta$. The environment moves to its next state through its state transition function, i.e. $s_{t+1}=\gT(s_t,a_t): \gS \times \gA \rightarrow \gS$, and provides an immediate and sparse reward $r_t$, defined as
\begin{equation}
    r_t = \gR(o_{t+1}, g) = 
    \begin{cases}
    0,      & \text{if } ||o_{t+1} - g||_2 \leq \epsilon \\
    -1,     & \text{otherwise}
\end{cases}
\end{equation}
where $\epsilon$ is a pre-defined threshold. Following its policy, the robot interacts with the environment until the episode terminates after $T$ steps. The interaction between the robot and the environment generates a trajectory, $\tau=(g,s_1,a_1,r_1,\hdots,s_T,a_T,r_T,s_{T+1})$. The ultimate learning objective is to find the optimal policy that maximises the expected sum of the discounted rewards over the time horizon $T$, i.e. 
\begin{equation}
\label{eq:rewards}
J(\mu_\theta) = \E_{\tau \sim  \sP(\tau|\mu_\theta)}[  R(\tau)=\sum_{i=1}^T{\gamma^{i-1}r_i}  ]
\end{equation}
where $\gamma$ is the discount factor. 

\subsection{Deep Deterministic Policy Gradient Algorithm}
\label{sec:ddpg}

Deep Deterministic Policy Gradient (DDPG) \citep{LillicrapHPHETS15} is adopted here as our main learning algorithm, however any other off-policy algorithm that operates on continuous action domains could be equally used. DDPG integrates deterministic policy functions \citep{SilverLHDWR14} with non-linear function approximators such as deep neural networks, and maintains a policy (actor) network $\mu_\theta(s_t,g)$ and an action-value (critic) network $Q^\mu(s_t,a_t,g)$. Further details about our implementation are provided in the Appendix.

\subsection{Hindsight Experience Replay}

Suppose  we are given an observed trajectory, $\tau=(g,s_1,a_1,r_1,\hdots,s_T,a_T,r_T,s_{T+1})$. Since $o_t$ can be obtained from $s_t$ using a fixed and known mapping, the path that was followed by the object during the trajectory, i.e. $o_1,\hdots,o_{T+1}$, can be easily  extracted. HER samples a new goal from this path, i.e. $\Tilde{g} \sim \{o_1,\hdots,o_T\}$, and the rewards are recomputed with respect to $\Tilde{g}$, i.e. $\Tilde{r}_t = \gR(o_{t+1}, \Tilde{g})$. Using these rewards and $\Tilde{g}$, a new trajectory is created implicitly, i.e.  $\Tilde{\tau}=(\Tilde{g},s_1,a_1,\Tilde{r}_1,\hdots,s_T,a_T,\Tilde{r}_T,s_{T+1})$. These HER trajectories $\Tilde{\tau}$ are used to train the policy parameters together with the original trajectories.

\section{Methodology}
\label{sec:methodology}

In this section, we explain the steps involved in our proposed procedure, i.e. (a) how object locomotion policies are learned, (b) how the novel curriculum learning model is defined upon them, and (c) how the  curriculum is implemented for robotic manipulation.

\subsection{Learning Object Locomotion Policies}
\label{sec:object_locomotion}

\begin{figure}[h!]
	\begin{center}
		\centerline{\includegraphics[width=\columnwidth,trim={0.2cm 7.2cm 0.6cm 2.4cm},clip]{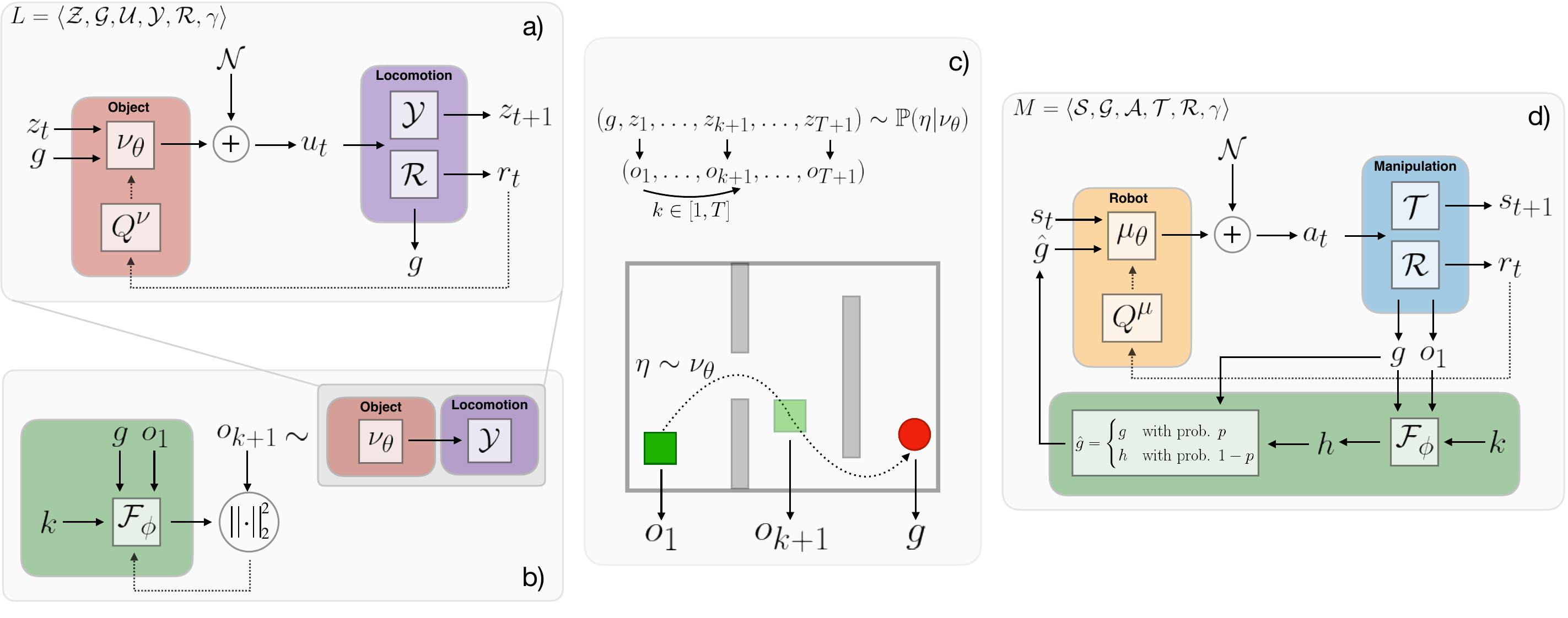}}
		\caption{a) Learning the object policy on locomotion. b) Adopting object policy to generate trajectories for training of $\gF_\phi$. c) A visual example for training of $\gF_\phi$: the green coloured object whose initial position denoted by $o_1$ needs to pass the obstacles and eventually reach the target position $g$ in red. $\nu_\theta$ is used to obtain the trajectory moving the object from $o_1$ to $g$. Using these trajectories, we train $\gF_\phi$ to predict the object position at time-step $k+1$ for given $o_1$ and $g$. d) $\gF_\phi$ is used to generate a curriculum over object positions on robotic manipulation. The difficulty of $h$ is adjusted through $k$ starting from $1$ and reaching $T$ as the robot performs better on imagined goals.
		}
		\label{fig:model}
	\end{center}
	\vskip -0.1 in
\end{figure}

The object involved in the manipulation task is initially modelled as an agent capable of independent decision making abilities, and its decision process is modelled by a separate MDP defined by a tuple $L=\langle \gZ, \gG, \gU, \gY, \gR, \gamma \rangle$. Here, $\gZ$ is the set of states, $\gG$ is the set of goals, $\gU$ is the set of actions, $\gY$ is the state transition function, $\gR$ is the reward function and $\gamma \in \left[0, 1\right)$ is the discounting factor. The same goal space, $\gG$, is used as in the robotic manipulation. $z_t \in \gZ$ is a reduced version of $s_t$ that only involves object-related features including the position of the object, i.e. $o_t \subset z_t$. The object's action space explicitly controls the pose of the object, and these actions are controlled by a deterministic policy, i.e. $u_t=\nu_\theta(z_t,g): \gZ \times \gG \rightarrow \gU$. The state transition is now defined on different spaces than robotic manipulation, i.e. $\gY: \gZ \times \gU \rightarrow \gZ$; however, the same sparse reward function is used here as before. Figure \ref{fig:model}a illustrates the training procedure used in this context and based on DDPG with HER. The optimal object policy $\nu_\theta$  maximises the expected return $J\left(\nu_\theta \right) = \E_{g,z_t \sim  \gD}[\sum_{i=1}^T{\gamma^{i-1}r_i}]$ where  $\gD$ denotes the replay buffer containing the trajectories, indicated by $\eta$, obtained by $\nu_\theta$ throughout training.

\subsection{Learning to Imagine Goals}

Upon training, the object locomotion policy $\nu_\theta$ can be used to generate trajectories, i.e. $(g,z_1,u_1,r_1,\hdots,z_T,u_T,r_T,z_{T+1}) \sim \sP(\eta|\nu_\theta)$. Since  $o_t \subset z_t$, an object path originating at $o_1$ and moving all the way to the final target $g$, i.e. $(o_1,\hdots,o_{T+1})$, can be obtained from $(z_1,\hdots,z_{T+1})$. We use these trajectories to train a model, $\gF_\phi$, to predict the object's position $k$-steps ahead within a trajectory. The model takes $o_1$ and $g$ together with an integer $k\in [1,T]$ as input, and outputs the prediction for the object position at time-step $k+1$, i.e. $\hat{o}_{k+1} = \gF_\phi(o_1,g,k)$. The model parameters $\phi$ are found such that (see also Figure \ref{fig:model}b):
\begin{equation}
    \argmin_\phi \EX_{\substack{(g,o_1,\hdots,o_T) \sim \sP(\eta|\nu_\theta) \\ 
                       \forall k \in [1,T]}}
                       \bigg[\Big|\Big|o_{k+1} - \gF_\phi(o_1,g,k)\Big|\Big|_2^2\bigg]
\end{equation}

\subsection{Learning the Robot Policy for Manipulation with Imagined Goals}

In this final step, the predictive model $\gF_\phi$ is leveraged to generate a sequence of imagined goals that can facilitate the learning of the original manipulation policy.  For every new episode characterised by $o_1$ and $g$, a predicted object position at time-step $k+1$, i.e. $h = \gF_\phi(o_1,g,k)$, is used as an intermediate goal to train the robot. By adjusting $k$ throughout the training, a  curriculum is formed.

\textbf{Adjusting the curriculum difficulty:} The value of $k$ controls the difficulty of an imagined goal $h$. Starting from $k=1$ and all the way up to $k=T$, one could simply increase $k$ by one every time the robot masters the current level. However, we have observed that this simple strategy can underperform due to some form of forgetting; this happens when the robot fails to master the current level, and the replay buffer starts to be filled with failing trajectories. Instead, $k$ is sampled from a uniform distribution, $U(1,k_{max})$, and $k_{max}$ is increased upon reaching fluency at the current difficulty level. This remedy resolves the forgetting issue since training at a given difficulty level now allows for goals at lower levels to be randomly introduced; this forces the robot to occasionally `practice' on simpler problems rather than always being confronted with more difficult ones. In order to decide whether the robot is fluent enough with the current difficulty level, we evaluate its performance for $k=k_{max}$, i.e. $h_{max}=\gF_\phi(o_1,g,k_{max})$. When the robot achieves a success rate higher than some threshold (e.g. $0.2$) for a given $h_{max}$, we increase $k_{max}$ by one, and keep increasing it until $T$ is reached. This scheme is somewhat inspired by the \textit{boundary-sampling} idea used in the Automatic Domain Randomization (ADR) algorithm \citep{akkaya2019solving}.  

\textbf{Learning the Robot Policy:} The policy $\mu_\theta$ is learned by DDPG as in Section \ref{sec:ddpg}. We ensure that both original goals $g$ and imagined goals $h$ are used during training. That is, goal $\hat{g}$ where $\hat{g}= g$ with probability $p$, and otherwise $\hat{g} = h$. Introducing $g$ early on in the training allows for faster learning as the policy may become capable to accomplish the original goals before the curriculum provided through $h$ is completed at $k=T$. Using a mixture of $g$ and $h$, the robot interacts with the environment and obtains trajectories $\tau=(\hat{g},s_1,a_1,\hat{r}_1,\hdots,s_T,a_T,\hat{r}_T,s_{T+1})$, where rewards are calculated with respect to $\hat{g}$, i.e. $\hat{r}_t=\gR(o_t,\hat{g})$. Experienced trajectories are stored into the replay buffer $\gD$. The parameters of $\mu_\theta$ are updated to maximise Eq. (\ref{eq:rewards}) using the trajectories sampled from $\gD$. Figure \ref{fig:model}d illustrates this procedure, and Algorithm \ref{algo:training_with_aux_goals} provides the pseudo code.

\setlength{\textfloatsep}{10pt}
\begin{algorithm}[t!]
	\label{algo:training_with_aux_goals}
	\begin{smaller}
		\DontPrintSemicolon
		\SetKwInOut{Given}{Given}
		\SetKwInOut{Initialise}{Initialise}
		\SetKwInOut{Return}{Return}
		\Given{Robotic Manipulation MDP $M=\langle \gS, \gG, \gA, \gT, \gR, \gamma \rangle$, 
		       Imagined goal generator $\gF_\phi(o_1,g,k)$, \\
		       Random process $\gN$ for exploration,
		       Fixed and known mapping function $m:\gS \rightarrow \gO$
		       }
		\Initialise{Parameters $\theta$ for robot policy $\mu_\theta$, Experience replay buffer $\gD$\\
		            }
        $k_{max} = 2$ \\
        \For{$\textit{episode} = 1$ \textbf{to} $N$}{
		    Receive initial state $s_1$ and $g$, $o_1=m(s_1)$\\
		    Sample $k \sim U(1,k_{max})$ where $U$ is uniform distribution\\
		    $h=\gF_\phi(o_1,g,k)$ \\
		    $\hat{g} = \begin{cases}
                                    g      & \text{with prob. } p \\
                                    h      & \text{with prob. } 1-p 
                        \end{cases} $
		    
		    \For{$t = 1, T$}{
		        Sample an action: $a_t = \mu_\theta\big(s_t,\hat{g})+\gN$ \\
		        Execute the action: $s_{t+1} = \gT(s_t,a_t)$ and $r_t = \gR(o_{t+1},\hat{g})$\\
		    }
		    Store $\tau=(\hat{g},s_1,a_1,\hat{r}_1,\hdots,s_T,a_T,\hat{r}_T,s_{T+1})$ in memory buffer $\gD$ \\ 

            Update $\mu_\theta$ with trajectories drawn from $\gD$ \\
            \If {$k_{max} \leq T$}{
                Test $\mu_\theta$ for $h_{max}=\gF_\phi(o_1,g,k_{max})$ \\
                \If {$success \geq threshold$}{
                $k_{max} = k_{max} + 1$
                }
            }
		}
		\caption{Learning manipulation policy with imagined goals}
	\end{smaller}
\end{algorithm}

\begin{figure}[b!]
	\begin{center}
		\centerline{\includegraphics[width=\columnwidth,trim={0.5cm 13cm 0cm 2.5cm},clip]{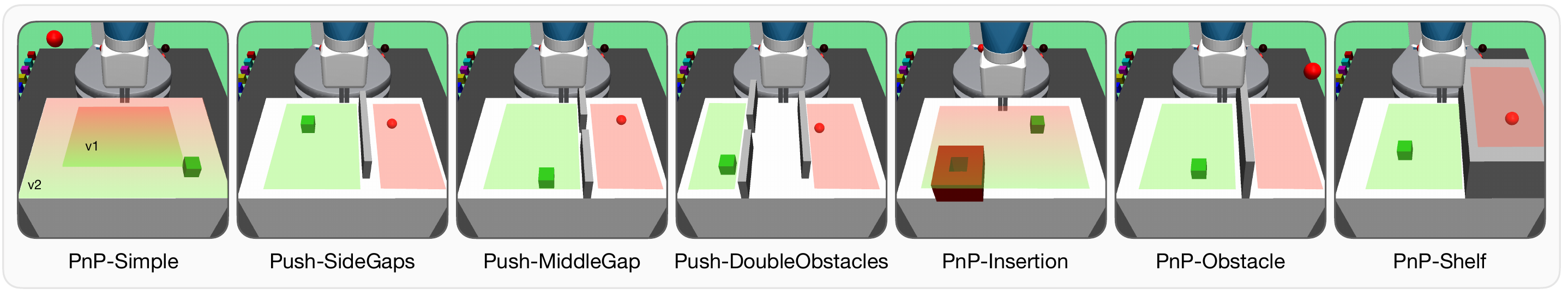}}
		\caption{Illustrations of the environments. The object is coloured in green and the target position is visualised as a red sphere. The green and red shaded areas indicate possible initial and target object positions.}
		\label{fig:env_vis}
	\end{center}
	\vskip -0.2 in
\end{figure}

\section{Experiments}
\label{sec:experiments}

\subsection{Environments}

We evaluate our method on 7 simulated MuJoCo environments that use a 7-DoF Fetch robotics arm. We build these environments upon the \textit{Push} and \textit{PickAndPlace} environments from \citep{PlappertHER2}. As visualised in Figure \ref{fig:env_vis}, the environments differ in terms of  initial and target object positions, as well as the obstacles between these two. In all the environments, the target indicates the desired 3D position of the object. A reward of $0$ is provided to the agent if the object is within 5-cm range to the target, and $-1$ otherwise. Robot actions are 4-dimensional: 3D to specify the desired arm movement in Cartesian coordinates and 1D to control the opening of the gripper. In \textit{Push} tasks, the robot is not allowed to control the opening of the gripper to prevent grasping. The observations include the positions and linear velocities of the robot arm and the gripper, as well as the object's position, rotation and angular velocity, and its relative position and linear velocity to the gripper. An episode terminates after $50$ time-steps, except for \textit{Push-DoubleObstacles} which terminates after $80$ time-steps. In the auxiliary object locomotion tasks, on the other hand, observations include the object's position, rotation and angular velocity. The object actions are 7 dimensional (3D for translation and 4D for rotation). In practice, we define the object as a \textit{mocap} MuJoCo entity which enables us to control its 7D pose. In all cases, the object actions correspond to the desired relative change in the object pose between two consecutive time-steps; however, their realisation depends on the dynamics executed by the simulator. The same reward function and termination criteria are used as in manipulation.  

\subsection{Implementation and Training Process}

We use three-layer neural networks with ReLU activations for all models, and optimise them using Adam \citep{KingmaB14}. We adopt the optimised hyperparameter values for HER (the baseline approach) from \citep{PlappertHER2}, and use them for all models. We train the models on a GPU enabled machine using its CPU cores to generate experiences and GPU to optimise the parameters. At each epoch of the training, we generate $38*50=1900$ full episodes and update the parameters 40 times using batches of size $4864$. We observed that the performance of the proposed approach is robust with respect to the model specific hyperparameters and report the results by choosing $p=0.2$ and a success threshold of $0.25$. An ablation study on the effects of these two hyperparameters can be found in Appendix, where we also describe all the remaining  hyperparameters in greater detail. 

\subsection{Comparison and Performance Evaluation}
\label{sec:evaluation}

We consider 6 alternative methods for comparison. All methods are built upon DDPG \citep{LillicrapHPHETS15} and implement HER \citep{AndrychowiczCRS17} as standard. Differences are as follows: \textit{HER} is the baseline as described in \citep{AndrychowiczCRS17} using sparse rewards. \textit{Shaped} uses distance-based shaped rewards instead of sparse rewards. \textit{SLDR} uses auxiliary rewards introduced by \citep{KilincSimulated2019} together with sparse rewards. In \textit{SLDR}, the optimal object locomotion actions are compared with those caused by the robot in terms of the action-values, and the difference is introduced as an auxiliary reward. \textit{RND} uses curiosity-based exploration bonuses together with sparse rewards. Exploration bonuses are obtained by adopting Random Network Distillation (RND) proposed in \citep{burda2018exploration}. Curriculum-guided HER (\textit{CHER}) \citep{fang2019curriculum} brings adaptive experience selection upon baseline HER. \textit{FO} refers to the approach introduced in this paper. We also combine our approach with SLDR and refer this combination as \textit{FO+SLDR}. Following \cite{PlappertHER2}, we evaluate the performances after each training epoch by computing the test success rate averaged across 380 deterministic rollouts with original goals $g$. In all cases, we repeat an experiment with 5 different random seeds and report results by computing the median test success rate with the interquartile range.

\begin{figure}[t!]
	\begin{center}
		\centerline{\includegraphics[width=\columnwidth,trim={0cm 0.2cm 0cm 0cm},clip]{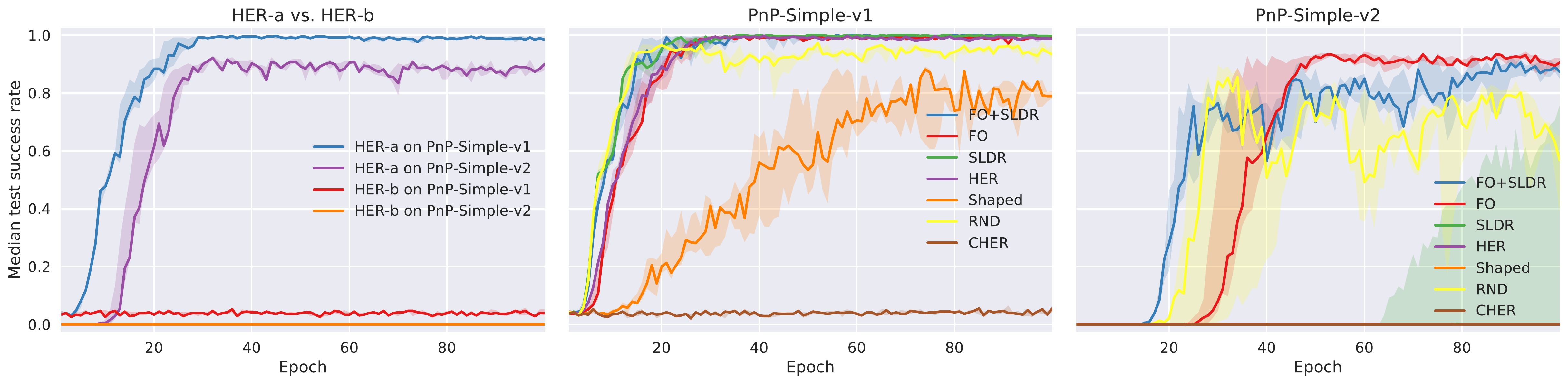}}
		\caption{a) The performances of HER-a and HER-b on both \textit{PnP-Simple} versions. b) The performances of all baselines on PnP-Simple-v1. c) The performances of all baselines on PnP-Simple-v2.}
		\label{fig:pnp_ablation}
	\end{center}
	\vskip -0.15in
\end{figure}

\textbf{Demonstrating the Problem on Simple \textit{PickAndPlace}:} To investigate how initial and target object positions affect learning, we define 2 different versions of \textit{PickAndPlace} (\textit{PnP}). \textit{PnP-Simple-v1} corresponds to the version mentioned in \citep{PlappertHER2}. In \textit{PnP-Simple-v2}, we simply remove two constraints that condition the initial and target object positions, i.e. unlike \citep{PlappertHER2} we do not require the positions to be sampled from within a smaller square centred around the gripper, and we do not enforce the targets to be on the table level in some of the training episodes. We adopt the baseline HER from \citep{AndrychowiczCRS17} to train two policies, i.e. \textit{HER-a} on \textit{PnP-Simple-v1} and \textit{HER-b} on \textit{PnP-Simple-v2}, following identical learning procedures. Throughout the training, we test their performances on both versions as presented in Figure \ref{fig:pnp_ablation}. \textit{HER-a} achieves \textit{PnP-Simple-v1} with $1.0$ success rate, and \textit{PnP-Simple-v2} with $0.9$. This result indicates  that the basic characteristics of the required robotic behaviours are similar in both versions, and \textit{PnP-Simple-v2} is not drastically harder to perform than \textit{PnP-Simple-v1}. However, \textit{HER-b} fails on both versions suggesting that \textit{PnP-Simple-v2} is harder to learn for HER. This is explained by the fact that easy-to-solve episodes are seen less frequently in \textit{PnP-Simple-v2} than in \textit{PnP-Simple-v1}. Figure \ref{fig:pnp_ablation} also demonstrates the performances of all 7 approaches on both versions. It can be noted that our approach is robust to such differences in the environments at a small cost of a slightly slower learning rate. We observe that this trade-off is generally compensated when combining \textit{FO} with \textit{SLDR}, as reported in the following section.

\begin{figure}[t!]
	\begin{center}
		\centerline{\includegraphics[width=\columnwidth,trim={0cm 0.2cm 0cm 0cm},clip]{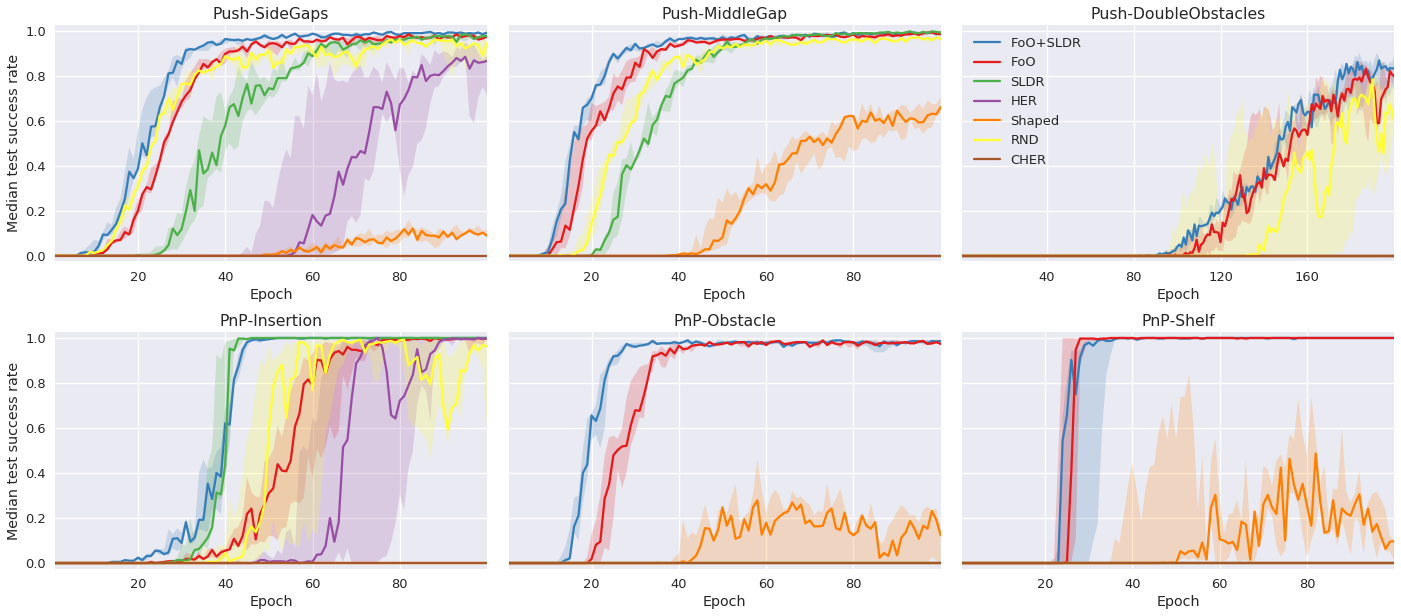}}
		\caption{The learning curves of 7 approaches on 6 environments. \textit{FO} can learn all 6 manipulation tasks with greater exploration demands. When combined with SLDR, \textit{FO+SLDR} results in faster learning compared to \textit{FO} alone.}
		\label{fig:push_obstacle}
	\end{center}
	\vskip -0.15in
\end{figure}

\textbf{Performance Evaluation on Other Environments:} Before evaluating the performances, it is worth noting a difference between \textit{Push} and \textit{PnP} tasks in terms of exploration requirements. \textit{Push} variations only require horizontal exploration in the space of object positions. Even a random robot policy may change the object position in the $x$- and $y$- axes, and contribute to horizontal exploration. \textit{PnP} variations, on the other hand, additionally require vertical exploration, which becomes possible only after the robot has learned to grasp the object. In general, the \textit{PnP} variations are associated with greater exploration demands. However, \textit{PnP-Insertion} proves to be easier than \textit{PnP-Obstacle} and \textit{PnP-Shelf}; unlike the other 5 environments, in this task the initial and target object positions are sampled from intersecting sets (see Figure \ref{fig:env_vis} for a visualisation.) 

The \textit{HER} baseline fails in most cases as the exploration becomes harder. Interestingly, in some cases, it is outperformed by \textit{Shaped}, which uses distance-based rewards as the only difference. \textit{RND} demonstrates that curiosity-based exploration bonuses can improve the performance beyond \textit{HER} as the exploration demand increases. However, it fails in more difficult cases such as \textit{PnP-Obstacle} and \textit{PnP-Shelf}. One can also notice the fluctuations in the learning curves of \textit{RND}. In terms of curiosity, the robot here would be rewarded when, for example, it throws the object out of the table for exploration. However, these rewards can jeopardise the learning process towards the main task. Although \textit{SLDR}'s manipulation-specific auxiliary rewards may seem as providing an advantage over \textit{RND}, \textit{SLDR} can also fail in more difficult environments, and in general its performance is poorer than \textit{RND}. We also observe that \textit{CHER} fails to learn our manipulation tasks. Overall, \textit{FO} outperforms these 5 approaches and achieves encouraging learning rates in all 6 environments. Moreover, the combined approach \textit{FO+SLDR} results in faster learning and has been proved to be the best approach overall on these environments. This is explained by the fact that both approaches, individually, motivate the robot to follow the object policy.

A potential limitation of the proposed approach is that not all the trajectories generated by the object locomotion policy are always guaranteed to be attainable by the robot. For example, in \textit{Push} tasks, occasionally the simulated object trajectories may pass over the obstacles, which clearly are not feasible for the robot. Our investigations have indicated that including unattainable imaginary goals do not hinder the performance of the proposed methodology in any significant way, and further discussions of these findings can be found in the Appendix.

\section{Conclusions}
\label{sec:conclusions}

We have proposed \textit{Follow the Object} (\textit{FO}), a new curriculum generation framework for learning robotic manipulation tasks with only sparse rewards. Using an object locomotion policy, a predictive model provides a curriculum of intermediary imagined goals resulting in high learning success rate. The object policies can be learned easily and realistically with a physics simulation engine. We have demonstrated that \textit{FO} is particularly beneficial in difficult manipulation problems requiring substantial exploration and can improve upon the performance obtained by existing algorithms. In future work, this framework can be extended to include more complex objects (e.g. flexible objects) and manipulation tasks, and account for uncertainty in the imagined goals, e.g. by using a distribution of predicted goals. 


\bibliographystyle{unsrtnat}
\bibliography{bibli}  

\end{document}